\documentclass{article}

\usepackage{microtype}
\usepackage{graphicx}
\usepackage{subfigure}
\usepackage{booktabs} 

\usepackage{hyperref}

\usepackage[utf8]{inputenc} 
\usepackage[T1]{fontenc}    
\usepackage{hyperref}       
\usepackage{url}            
\usepackage{booktabs}       
\usepackage{amsfonts}       
\usepackage{nicefrac}       
\usepackage{microtype}      
\usepackage{graphicx}
\usepackage{amsthm}
\usepackage{amsmath}
\usepackage[english]{babel}
\usepackage[T1]{fontenc}
\usepackage[utf8]{inputenc}
\usepackage{authblk}
\usepackage{bbm}
\usepackage{wrapfig}

\usepackage{adjustbox}

\usepackage{hyperref}
\usepackage{url}
\usepackage{soul}
\usepackage{ulem}
\usepackage{colortbl}
\usepackage[table]{xcolor}

\newtheorem*{theorem*}{Theorem}

\newtheorem*{prop*}{Proposition}

\usepackage[disable]{todonotes}

\newcommand{\chugo}[1]{\todo[color=red!60,inline]{H:#1}}

\usepackage[nonatbib, final]{corl_2021}

\author[1]{Hugo Caselles-Dupr\'e}

\author[2]{Michael Garcia-Ortiz}

\author[1]{David Filliat}

\affil[1]{U2IS, ENSTA Paris, Institut Polytechnique de Paris \& INRIA Flowers}
\affil[2]{CitAI, SMCSE, City University of London}

\title{SCOD: Active Object Detection for Embodied Agents using Sensory Commutativity of Action Sequences}

\begin{document}




\maketitle

\begin{abstract}

We introduce SCOD (Sensory Commutativity Object Detection), an active method for movable and immovable object detection. SCOD exploits the commutative properties of action sequences, in the scenario of an embodied agent equipped with first-person sensors and a continuous motor space with multiple degrees of freedom. SCOD is based on playing an action sequence in two different orders from the same starting point and comparing the two final observations obtained after each sequence. Our experiments on 3D realistic robotic setups (iGibson) demonstrate the accuracy of SCOD and its generalization to unseen environments and objects. We also successfully apply SCOD on a real robot to further illustrate its generalization properties. With SCOD, we aim at providing a novel way of approaching the problem of object discovery in the context of a naive embodied agent. We provide code and a supplementary video\footnote{\url{https://youtu.be/Bc5fwZH-CQU}}.

\end{abstract}

\section{Introduction}

\chugo{state-of-the-art research in baby/children's moveable/immoveable object detection ability, improve immovable object detection}

The role of active movement in object discovery is crucial in the cognitive development of children \cite{colombo2001development, johnson2010infants} and and is considered a key aspect in theories of perception \cite{gibson2014ecological}. Children and animals gradually construct their perception of the environment, relying on basic mechanisms such as eyes movement and motor babbling \cite{baillargeon1985object}. By actively interacting with its environment and observing the resulting changes in sensory stimuli, the subject gradually learns about the boundaries, objects, and free spaces of its environment. 

\begin{figure*}
    \centering
    \includegraphics[scale=0.7]{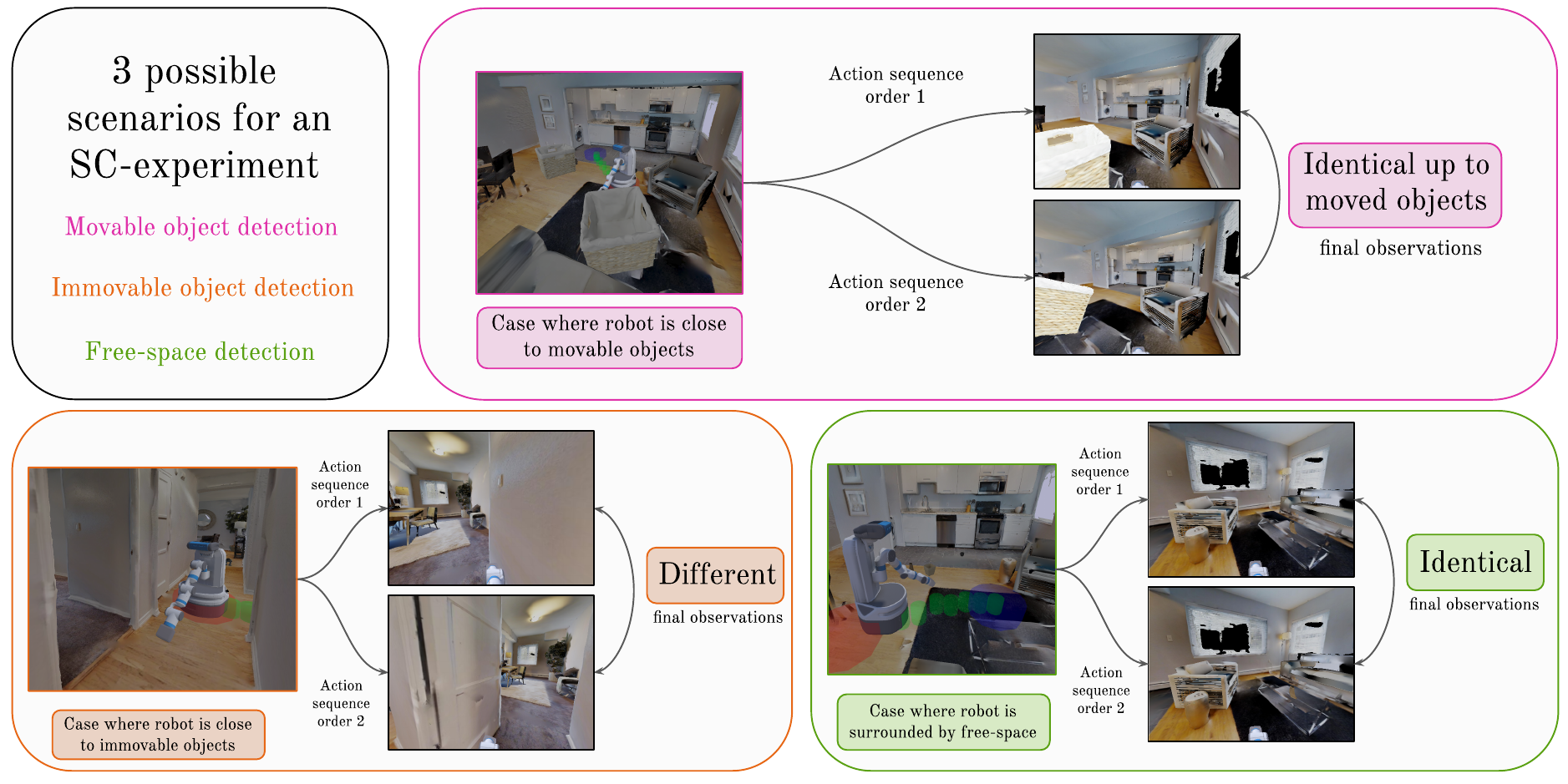}
    \caption{Intuition for our approach SCOD for object discovery. 3 scenarios are possible after a SC-experiment (i.e. playing an action sequence in two different orders from the same starting point), depending on the surroundings of the agent. If the agent is surrounded by free-space then the two final observations $obs_1$ and $obs_2$ will be identical, whereas if it is surrounded by immovable objects (walls, sofa) then $obs_1$ and $obs_2$ will probably be different (because of the different interactions with the immovable objects). If the agent is surrounded by movable objects, $obs_1$ and $obs_2$ will be identical up to moved objects that have been interacted with in different manner in the two action sequences.}
    \label{fig:scod_intuition}
\end{figure*}

The learning scenario we consider consists in embodied agents situated in realistic environments, i.e. agents that face partial observability, coherent physics, first-person view with high-dimensional state space, and low-level continuous motor (i.e. action) space with multiple degrees of freedom. These embodied agents, act in such environments by producing a stream of sensorimotor data, composed of successions of motor states and sensory information. The agent actively moves in the environment to manipulate this stream of data and learn about its surroundings. 

The present work introduces an object discovery method for such embodied agents termed SCOD (Sensory Commutativity Object Discovery), based on movement and the sensory commutativity properties of action sequences. Starting from the sensory commutativity experiment (SC-experiment) \cite{caselles2020sensory}, which consists in playing an action sequence in two different orders from the same starting point, we aim at extracting information about what the agent can interact with in the environment. 

We assess commutativity properties by comparing the two final observations $(obs_1, obs_2)$ obtained after each sequence. Based on previous work that study sensory commutativity \cite{philipona2004perception, philipona2008developpement}, we posit that there are three potential outcomes when comparing $obs_1$ and $obs_2$: they are either completely different, identical, or identical up to moved objects. We thus provide the agent with a basic ability: being able to compare two images. The agent acquires this skill in a pre-training phase where a mask predictor is learned, which architecture is based on optical flow prediction. Studies in cognitive science indicate that children are capable of doing this differentiation at a very young age (1 month old) \cite{kaufmann1995development, johnson2010infants}, so equipping the naive agent with this basic ability is a reasonable assumption. The mask predictor takes two images as input, and outputs two masks corresponding to what has moved between the two observations. Combining the mask predictor and the SC-experiments allows the agents to discover immovable and movable objects in its surroundings. The intuition behind the approach are presented in Fig.\ref{fig:scod_intuition} and on the supplementary video\footnotemark[1].

We use the iGibson interactive environment \cite{shenigibson} for simulating this embodied agent scenario. We control the 10-DOF robot Fetch \cite{wise2016fetch}, in iGibson's 3D scenes reconstructed from real homes. Our results qualitatively and quantitatively showcase the accuracy and generalization properties of SCOD. We also provide a review on object detection and discovery, and compare our work to current state-of-the-art object detection methods, video object segmentation and tracking methods and methods from the robotics literature. 

Our contributions are the following:

\begin{itemize}
    \item We propose SCOD, a novel object discovery method based on sensory commutativity of action sequences.
    \item We demonstrate the accuracy and generalization properties of SCOD on 3D realistic robotic setups by using the Fetch robot in the iGibson interactive simulator. We also provide real-life generalization examples.
    \item We compare SCOD to the current landscape of object detection methods. We provide analysis and comparisons to better understand how novel the approach is.
\end{itemize}

\section{SCOD: Object Discovery using Sensory Commutativity}

\begin{figure*}
    \centering
    \includegraphics[scale=0.7]{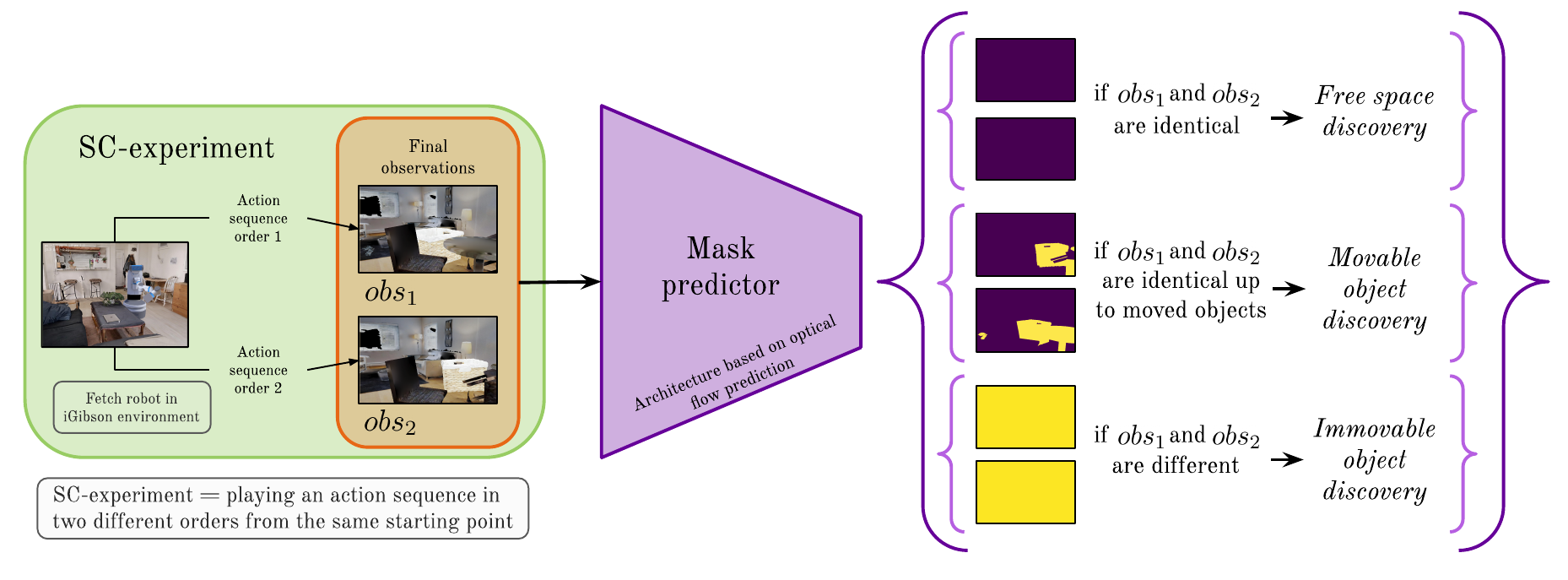}
    \caption{Overview of our approach SCOD for object discovery. The agent plays an action sequence in two different order from the same starting point (SC-experiment). From the two resulting final observations $(obs_1, obs_2)$, we aim at learning about objects in the surroundings of the agent, see Fig.\ref{fig:scod_intuition} for the reasoning behind this idea. For that, a mask predictor is pretrained on procedurally generated data, and applied on $(obs_1, obs_2)$. The mask predictor outputs two binary "difference" masks that represent the semantic difference between $(obs_1, obs_2)$: either identical (all zeros), completely different (all ones) or identical up to moved objects (segmentation masks).}
    \label{fig:scod_overview}
\end{figure*}

\subsection{Background on sensory commutativity}

Our work is based on the concept of sensory commutativity, which studies the commutative properties of action sequences with respect to sensory information. It is concerned with the questions of how an embodied agent learns the properties of its environment and its own body. The theory has a proper theoretical grounding \cite{philipona2008developpement, caselles2020sensory} and has applications such as understanding the concept and dimensionality of space \cite{philipona2003there} or gaining information about the body of the agent \cite{caselles2020sensory}. 

SCOD is based on the properties of sensory commutativity, learned through Sensory Commutativity-experiment (SC-experiment), which consists in having the agent play an action sequence in two different orders from the same starting point and comparing the two outcomes in observations (are the observations identical or not?). SC-experiments are usually the basis for studying sensory commutativity.

\subsection{Motivation}

We propose to rely on an analysis of the differences between the two observations $obs_{1}$ and $obs_{2}$ resulting from an SC-experiment.

We posit that comparing $obs_{1}$ and $obs_{2}$ leads to three possible outcomes from which the agent can learn about immovable and movable objects in the environment (these scenarios are illustrated in Fig.~\ref{fig:scod_intuition}):

\vspace{0.2cm}

\begin{itemize}
    \item $obs_{1}$ and $obs_{2}$ are different: the two action sequences from this starting position do not commute, because the robot interacted with immovable objects. Consider for instance the robot in a still stance with its arm straight such that the robot stand with a wall at its right. Rotating the base of the robot to the left then to the right would end up with observation $obs_1$, which is the same observation as the starting situation. Now if the robot rotates its base right then left, since it's blocked by the wall from trying to right first, the robot will end up left to where it started, and it will observe $obs_2 \neq obs_1$. Using the position of the agent, we can now map immovable objects in the environment.
    \item $obs_{1}$ and $obs_{2}$ are identical: the two action sequences from this starting position commute, because the robot did not interact with anything in the environment (free movement). An example would be the same situation as in the last paragraph, but with a starting position where the robot is not next to a wall, and stands in a place where there is free space. Rotating left then right, or right then left yields the same observations $obs_1 = obs_2$. Using the position of the agent, we know that there are no objects in the current space around the robot.
    \item $obs_{1}$ and $obs_{2}$ are identical except for an object that has been moved: it's the case where the robot has interacted with a movable object that did not block the robot's movement. An example would be having the robot with a movable object to its right in its sight. Rotating left then right would leave no changes in observations, while rotating right then left would push the object out of its sight. Hence the two action sequences did commute, except for the object that has been moved. We can learn to detect this moving object and track it.
\end{itemize}

We therefore posit that from these outcomes, the robot can discover and map immovable and movable objects in the environment.

\subsection{Object discovery method}
\label{sec:proc_data}

In order to verify the aforementioned intuition, the robot needs to be able to perform an SC-experiment and then detect: 1) if the two resulting observations are identical or not, 2) if they are identical except for the parts of an image corresponding to an object that moved. For that, we equip the agent with a vision system that gets two observations as input and outputs two masks which will be all zeros if the two observations are identical, all ones if they are different, and the mask of an object if this object moves. Children are capable of this at a very young age so we consider this ability as innate and not acquired \cite{kaufmann1995development, johnson2010infants}.

\quad \textbf{Mask predictor training.} We thus train a neural network (whose architecture is discussed in the next section) with generated data to predict those two masks with two observations as input. We refer to this model as the "mask predictor". The data to train this model is collected by starting at a random position in the environment (observation $obs_1$) and then collecting data for the three possible outcomes. 
\begin{itemize}
\item no difference: it suffices to keep the same observation and the corresponding masks are all zeros. The data is ($obs_1$ + all zeros mask, $obs_1$ + all zeros mask).
\item completely different: we move the robot and get a different observation $obs_2$, the corresponding masks are all ones. The data is ($obs_1$ + all ones mask, $obs_2$ + all ones mask).
\item no difference except moved objects: we randomly disturb the orientation and position of some movable objects and get a new observation $obs_2$ identical to $obs_1$ up the moved objects. The data is ($obs_1$ + moving objects mask, $obs_2$ + moving objects mask)
\end{itemize}

\begin{figure}[h!]
    \centering
    \includegraphics[scale=0.15]{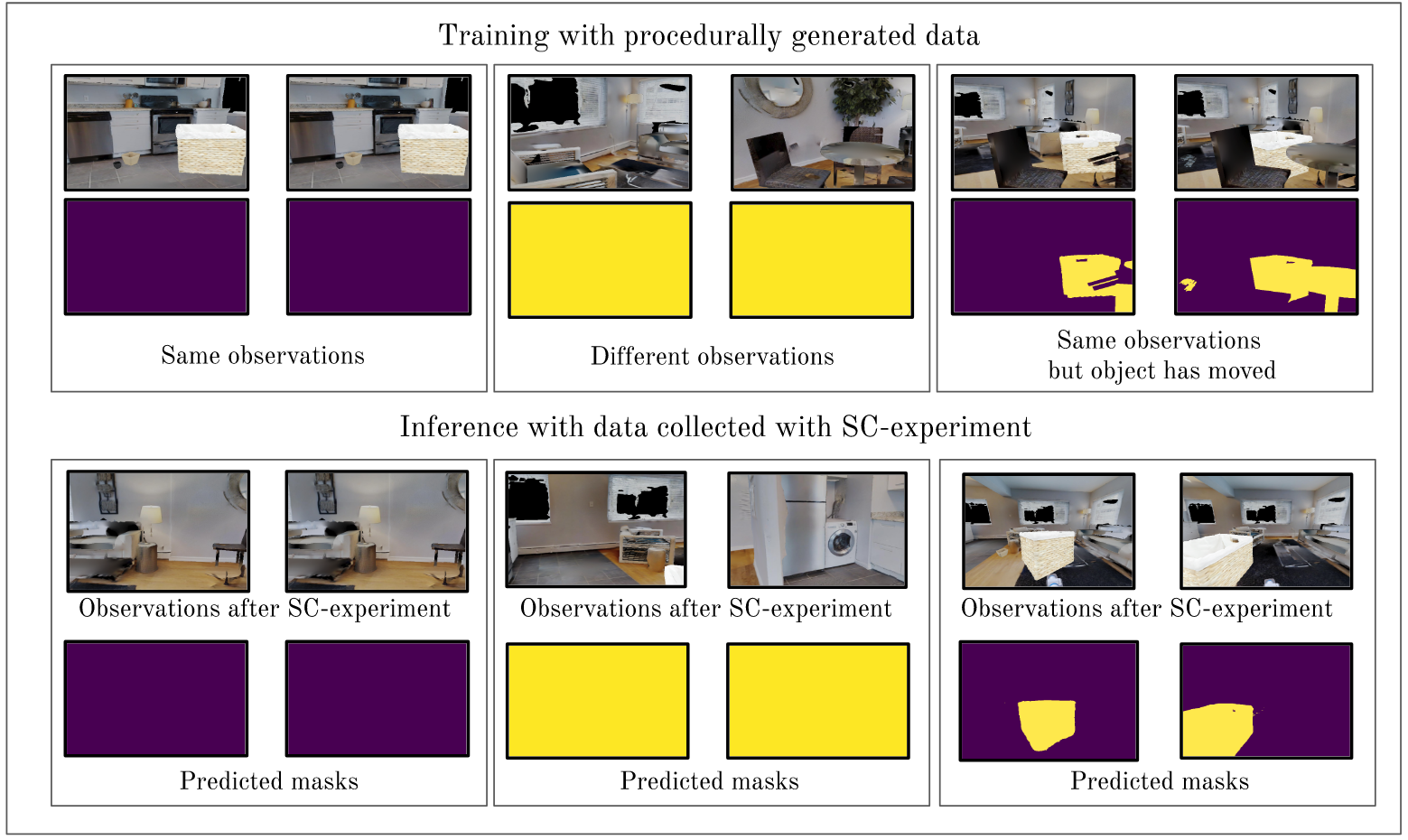}
    \caption{\textbf{Top: }dataset for training the mask predictor. \textbf{Bottom: }inference results on data collected with SC-experiments (each image is the result of one action sequence). The dataset is procedurally generated to simulate the three possible scenarii resulting from a SC-experiment. \textbf{Left:} scenario where there are no changes in the observations. \textbf{Middle:} scenario where the observations are different. \textbf{Right:} scenario where the observations are identical up to moved objects.}
    \label{fig:dataset_obj}
\end{figure}

The resulting dataset is illustrated in Fig.\ref{fig:dataset_obj}.

\quad \textbf{Object discovery inference.} Once the mask predictor is trained, we place the agent in a random position in the environment and perform SC-experiments where we let it play an action sequence in different orders from the same starting point. Then, the goal is for the agent to detect immovable and movable objects using the generated data from the SC-experiments and the mask predictor. 

To summarize, our object discovery method SCOD is divided in two steps:

\begin{itemize}
    \item Step 1: Train the mask predictor on procedurally generated data.
    \item Step 2: Discover movable and immovable objects by letting the agent perform SC-experiments and use the trained mask predictor on the resulting observations.
\end{itemize}

\begin{figure*}[t!]
    \centering
    \includegraphics[scale=0.5]{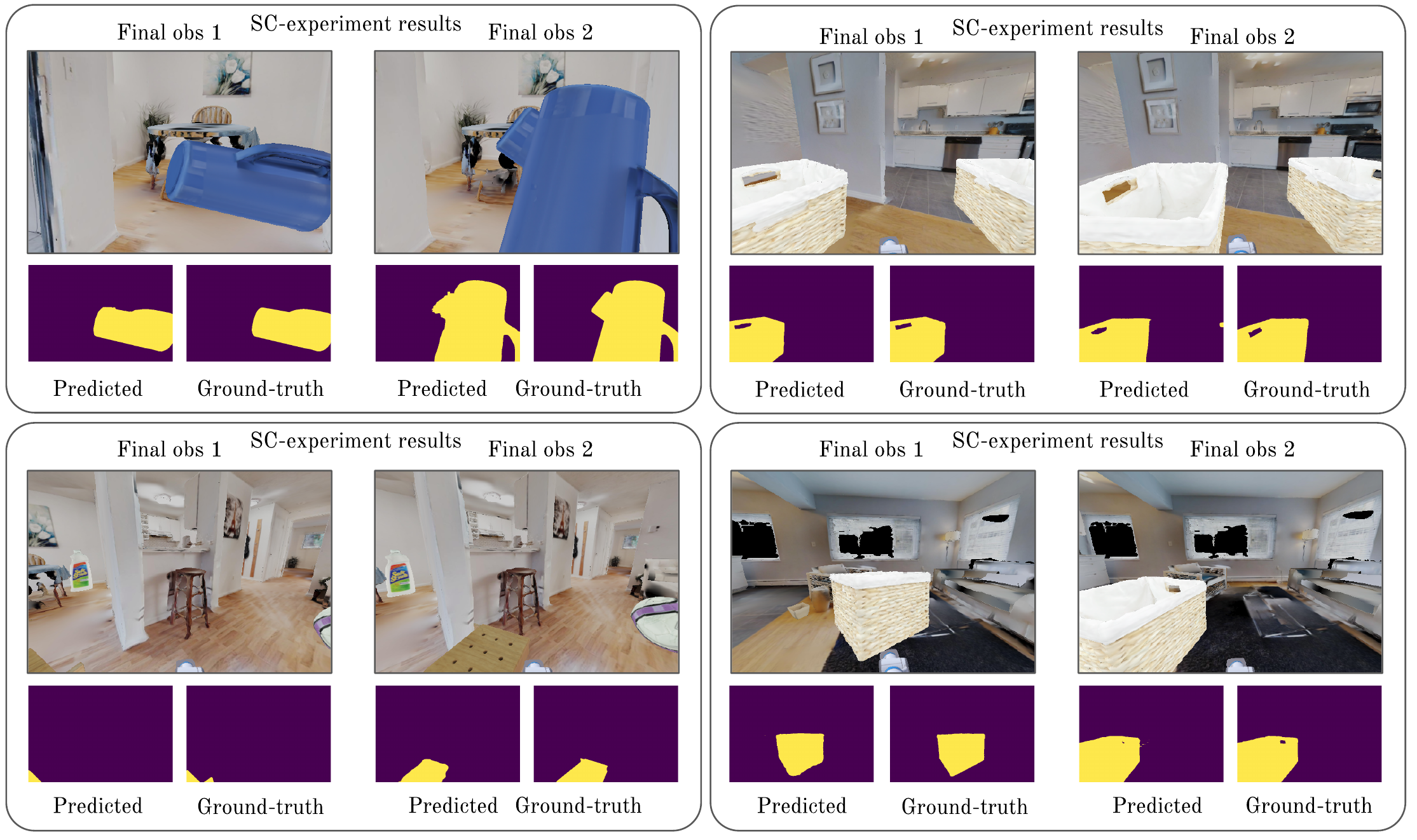}
    \caption{Generalization study of movable object detection using SCOD with a mask predictor trained in the Placida environment. In all scenarios, our method correctly predicts the mask object. \textbf{Upper left:} Object and environment not seen during training. \textbf{Lower left:} Object, environment and field of view not seen during training. \textbf{Upper and lower right:} Field of view not seen during training.}
    \label{fig:obj_more}
\end{figure*}

\section{Experimental setup}

\subsection{Simulation and environment}

We use the iGibson interactive environment \cite{shenigibson} for simulating this embodied agent scenario. This simulation has various realistic indoor environments captured inside human habitations and numerous objects to interact with. In our experiments, we control Fetch, a 10-DOF real robot \cite{wise2016fetch} equipped with a 7-DOF articulated arm, a base with two wheels, a liftable torso and a RGB camera in its eye.
  
\subsection{Mask predictor training (step 1)}

\quad \textbf{Architecture choice reasoning.} Predicting the masks given the observations is a process similar to predicting the optical flow of two consecutive frames in a video. In this problem, the state-of-the-art neural network architecture predict the optical flow field ($2*W*H$) using two consecutive RGB frames of a video as input ($2*3*W*H$). The optical flow field is a projection of the motion field, i.e. the real world 3D motion between the two frames. Thus, the optical flow field corresponds to the displacement of pixels between the two frames. This type of architecture is adapted to our problem since we aim at predicting two binary masks ($2*W*H$), using the two final observations $(obs_1, obs_2)$ from the SC-experiment as input ($2*3*W*H$). The mask predictor estimates the displacement of objects between the two frames, a similar goal as in optical flow prediction.

For selecting the architecture, we first tested the FlowNet-S architecture \cite{fischer2015flownet}, a popular baseline for optical flow prediction, as a proof of concept. We then adopted the state-of-the-art RAFT model \cite{teed2020raft}, which performed better. We provide comparisons of the two model performances in the experiments.

\quad \textbf{Datasets.} For training the mask predictor, we procedurally generate 40k training data in the format of tuples $(obs_1, obs_2, mask_1, mask_2)$, as described in Sec.\ref{sec:proc_data}. We use the Placida environment, augmented with 40 objects from the YCB object benchmark \cite{calli2015benchmarking}. 

\quad \textbf{Training.}  For both the Flownet and RAFT model, we train the models using the same architecture and optimization process as proposed by their authors, except for the loss function and the output activation function. We change the loss function to a binary cross-entropy loss between the ground truth mask and the output mask of the network. We select the sigmoid function as output activation function so that the model outputs binary masks instead of the original optical flow map output ($2*W*H$). All training details are available in the original open-source implementations we used \footnote{\href{https://github.com/princeton-vl/RAFT}{Link to RAFT} and \href{https://github.com/ClementPinard/FlowNetPytorch}{link to FlowNet} implementations}.

\subsection{Object discovery using SC-experiments (step 2)}

\quad \textbf{SC-experiments.} The second part of our object discovery pipeline is to compute SC-experiments and use the trained mask predictor. For the SC-experiments, we play an action sequence in two different orders from the same starting point. We do this by resetting the environment between two action sequences. 

To illustrate our results, the action sequences we consider are composed of random (sign, amplitude) motor commands for the DOF of the arm that is closest to the body of the agent. Each action is applied for $\frac{1}{10}$th of a second, and the length of action sequences is set to $20$. Note that the choice of DOF is arbitrary, and any other DOF would have worked also. Yet, for illustrative purposes, this DOF empirically allows to have a well balanced mix of all the three possible scenarios of the SC-experiments.

We test our object discovery method in the Placida environment, as well as different environments not seen during training, such as the Bolton or RS environment. Similarly, in the object discovery step, the objects to detect are not necessarily seen during training, and come from the YCB object benchmark \cite{calli2015benchmarking}.

\quad \textbf{Evaluation.} For qualitative and quantitative evaluation, we manually create a test set with $50$ tuples $(obs_1, obs_2, mask_1, mask_2)$ of the three possible scenario resulting from a SC-experiment. We cannot construct this dataset automatically, as the mask has to be manually created by either assessing if the two observations are different or identifying which object has moved between the two observations. Using this dataset, we can first assess the prediction accuracy among the three possible outcomes of a SC-experiment. 

In the case where an object has moved (see example in lower right corner of Fig.\ref{fig:dataset_obj}), we can further analyze the accuracy of the predicted mask using the Jaccard index, or Intersection over Union ($IoU$), which is usually used in object segmentation literature \cite{athar2020stem}. It quantifies the overlap between predicted $(p1, p2)$ and ground-truth $(gt_1, gt_2)$ masks. It is defined as $IoU_i = \frac{|p_i \cap gt_i|}{| p_i \cup gt_i|}$.

\section{Results}

Our results are best illustrated in video, which we provide as supplementary material\footnote{\url{https://youtu.be/Bc5fwZH-CQU}}. We now present a qualitative and quantitative evaluation of the results.

\begin{table*}[h!]
\caption{Quantitative results for movable objects, immovable objects and free space detection using Flownet and RAFT architecture for the mask predictor. Both methods are tested on a test set for training, and on a test set for generalization to unseen environments and objects.}
\label{tab:quanti-res}
\begin{center}
\begin{sc}
\resizebox{\textwidth}{!}{\begin{tabular}{l|ccc|cccr}
\toprule
 & & No generalization test set & & & Generalization test set & \\
\midrule
Mask predictor & Movable object & Immovable object & Free space & Movable object & Immovable object & Free space \\ 
\midrule
Flownet & 0.86 (IoU) & 61.9\% (accuracy) & 95.8\% (accuracy) & 0.61 (IoU) & 75.0\% (accuracy) & 99.3\% (accuracy)  \\
RAFT & \textbf{0.97} (IoU) & \textbf{90.4\%} (accuracy) & 96.1\% (accuracy) & \textbf{0.84} (IoU) & \textbf{90.9\%} (accuracy) & 98.6\% (accuracy) \\

\bottomrule
\end{tabular}}
\end{sc}
\end{center}
\vskip -0.1in
\end{table*}

\subsection{Quantitative results}

We present the quantitative results on the manually collected test set in Tab.\ref{tab:quanti-res} (see non-generalization test set column). With the RAFT architecture for the mask predictor, we reach an average Jaccard index of $0.97$ for movable object detection and respectively $90.4\%$ and $96.1\%$ for immovable object and freespace detection. These results highlights the efficiency of SCOD for all three tasks.

Regarding the choice of mask predictor architecture, RAFT overall performance is significantly better than than FlowNet. This justifies the adoption of RAFT for the mask predictor. We also provide qualitative comparison of RAFT vs. Flownet predictions in App.\ref{app:raft_vs_flownet} which further highlights the superiority of RAFT in the accuracy of prediction.

\subsection{Qualitative results}

\quad \textbf{Can SCOD detect movable objects?} Results presented in Fig.\ref{fig:dataset_obj} \& \ref{fig:obj_more} illustrate that using the mask detector with the outcome of these SC-experiments allows to detect objects that have been moved. Note that the mask detector only detects objects that have moved between the two resulting observations, rightfully ignoring the other potential objects that were not moved.

\quad \textbf{Object tracking after detection.} After this detection, we can then use semi-supervised tracking algorithms in order to track the detected object. The results are illustrated in App.\ref{app:obj_tracking} and in the supplementary video\footnotemark[3]. These methods takes as input a video and the object mask of the first frame. We predict the first mask using SCOD, and then track the detected object using Space-Time memory networks \cite{oh2019video}. We obtain a quasi-perfect  automatic tracking of the detected object.

\begin{wrapfigure}{r}{0.4\textwidth}
    \centering
    \includegraphics[scale=0.5]{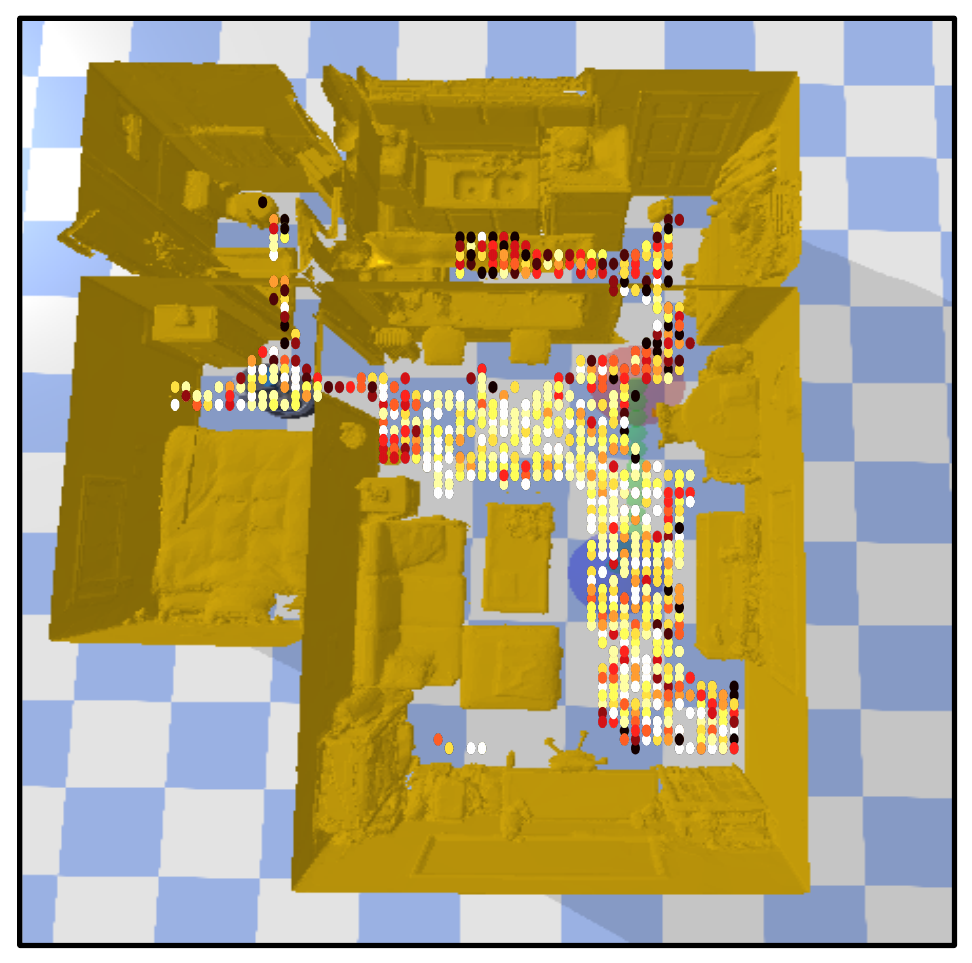}
    \caption{Immovable object detection using SCOD. Each dot represent the probability of observing the outcome "different" when playing a SC-experiment at this position (darker is higher). Free spaces are filled with white dots and cramped spaces with darker dots.}
    \label{fig:scp_map}
\end{wrapfigure}

\quad \textbf{Can SCOD detect immovable objects and free spaces?} Results presented in Fig.\ref{fig:dataset_obj} illustrate that the mask predictor is also able to accurately predict when the observations are different or identical. By isolating those two cases from the case where only one or a few objects have moved, we can map the starting position of the agent with the probability that an SC-experiment will commute. In Fig.\ref{fig:scp_map}, each dot represent a starting position, and the dot's color is the probability of observing the outcome "different" when playing a SC-experiment at this position (darker is higher). Regions with dark dots correspond to regions where there are walls and immovable objects in the way of Fetch's arm, whereas regions with white dots correspond to free spaces.

Indeed, in the kitchen part (room at the top), the space is cramped and so most of the positions indicate low commutation probability (less than $0.4$) because of the interactions induced with the furniture. In the living room (main room) and the bedroom (at the left), most empty space show high probability (around $0.8$ and $1.0$). We thus obtain a mapping of immovable objects and free spaces using SCOD predictions.

\subsection{Generalization study} 

\quad \textbf{Does SCOD generalize to unseen environments and objects?} In principle, SCOD only relies on having a precise mask predictor, which could thus work in any environment, any objects and any field of view. We thus performed a generalization study of our method. We manually created a generalization test set (150 instances) with data consisting of objects, environments and field of view that were not shown during training. For this study, we selected the Bolton environment, 20 objects from the YCB benchmark that were not shown during training, and a bigger field of view (90 versus 45 for training). In Fig.\ref{fig:obj_more} and Tab.\ref{tab:quanti-res}, we show results for the generalization study, which indicate that the mask predictor can indeed be used with environments, objects, and field of view that have been not shown during training. 

Qualitatively, the mask predictor is able to precisely predict which objects has moved, regardless of the shape of the object, and of the nature of the background and field of view. Quantitatively, the precision of SCOD shows strong generalization. We reach an average Jaccard index of $0.84$ for movable object detection. For immovable object detection and freespace detection, the performance is similar between the generalization and non-generalization test set. Hence, the low performance drop between in-distribution and out-of-distribution test sets allows us to conclude that SCOD generalizes to new environments, objects and field of view.

\quad \textbf{Real robot generalization.} As a last generalization test, we performed SC-experiments in real-life by using a Turtlebot robot. Qualitative results are presented on Fig.~\ref{fig:real-life} in Appendix \ref{app:real-life} and illustrated on the supplementary video\footnotemark[3]. We use the same mask predictor, which has been trained on synthetic images solely, and we obtain satisfying qualitative results. The methods seems able to bridge the reality gap. 

\section{Related work}
\label{sec:related_work}

\subsection{Passive object detection}

\quad \textbf{Object detection on still images.} Passive object detection method rely on largely annotated databases of object classes that allow to train fast and accurate predictive models for bounding boxes and segmentation masks. Pascal VOC \cite{everingham2015pascal} and COCO \cite{lin2014microsoft} are examples of the most used datasets, and popular methods like Mask R-CNN \cite{he2017mask} or YOLO \cite{redmon2016you} and their variants have shown to be very efficient at solving object detection and segmentation tasks. 

These models are by far the most used in practice: they are extremely useful for a number of real-life applications. Yet, these methods are not suited for open-ended object detection for an agent. The methods are not object-agnostic: they detect objects based on their similarity to objects seen during training. Objects exists in various shapes, colors and sizes in open-world exploration, and while a sufficiently large datasets might do the trick, this design does not seem suited for this type of experience. 

\subsection{Active object detection}

In robotics, there is a large body of work on active object detection (AOD) \cite{bohg2017interactive}, which is closer to SCOD. The goal is generally to perform active movements, such as poking and pushing in order to learn about movable objects. AOD methods are either based on fix viewpoint or first-person viewpoint. 

\quad \textbf{AOD with fixed viewpoint.} In the fix viewpoint scenario, we usually find a robotic arm (without a body or head) in front of a table with multiple objects to interact with \cite{schiebener2011segmentation,gupta2012using, lyubova2016passive, goff2019building, eitel2020learning}. This particular setup is the most common in AOD, as it has many real-life applications (in logistics for instance). However, it comes with constrains that do not apply to embodied agents, who face first-person viewpoints with partial observability of scenes. 

\quad \textbf{AOD with first-person viewpoint.} This setup is the closest to ours. The most common strategy is for the agent to push an object and then update its knowledge of movable objects \cite{bersch2012segmentation,schiebener2014physical}. This movement is complex for a robot with multiple DOF, which is why this movement is pre-programmed in those methods. In SCOD, we propose an alternative that only relies on random movements and do not make assumptions about available manipulation movements. Also, some methods do not use RGB as input for the object detection methods, but instead 3D cameras \cite{xu2015autoscanning}. This enhanced input allows to perform object detection in real time, which might be useful for application but does not resolve the problem of learning visual perception from RGB images.

\section{Discussion and conclusion}

\quad \textbf{Limitations.} We deployed SC-experiments in real-life with Turtlebot as a demonstration, but there are a few difficulties for a more advanced real-life deployment of SCOD. We need the agent to play two action sequences from the same starting point. In real-life, the method has to overcome stochasticity and irreversible actions (e.g. breaking a glass) which break that assumption. Also, if an object is moved, you would have to place it back to its original position. 
However, this could be overcomed by learning an accurate forward model of the environment that allows the agent to predict what will happen when it plays an action sequence. The forward model would act as a proxy for one of the sequence, and the robot would perform the other sequence in real life, therefore performing SC-experiments by comparing real experience with imagination. Recent works have made significant progress in this direction \cite{ha2018recurrent, hafner2020mastering}. We believe this is an important future work for using sensory commutativity to build perception for artificial agents. Finally, we also discuss in Appendix \ref{app:design} the algorithm design alternatives for the SCOD algorithm.




\quad \textbf{Conclusion.} We proposed SCOD, an object detection method based on sensory commutativity of action sequences. By playing an action sequence in two different orders from the same starting point, and comparing the two resulting observations using a mask predictor, the agent can learn a segmentation mask of movable objects, and detect walls, immovable objects and free spaces. SCOD departs from usual passive object detection methods, and provide a novel approach for active object detection in embodied scenarios.

\bibliography{bibli}

\begin{thebibliography}{10}

\bibitem{athar2020stem}
Ali Athar, Sabarinath Mahadevan, Aljo{\v{s}}a O{\v{s}}ep, Laura Leal-Taix{\'e},
  and Bastian Leibe.
\newblock Stem-seg: Spatio-temporal embeddings for instance segmentation in
  videos.
\newblock {\em arXiv preprint arXiv:2003.08429}, 2020.

\bibitem{baillargeon1985object}
Renee Baillargeon, Elizabeth~S Spelke, and Stanley Wasserman.
\newblock Object permanence in five-month-old infants.
\newblock {\em Cognition}, 20(3):191--208, 1985.

\bibitem{bersch2012segmentation}
Christian Bersch, Dejan Pangercic, Sarah Osentoski, Karol Hausman, Zoltan-Csaba
  Marton, Ryohei Ueda, Kei Okada, and Michael Beetz.
\newblock Segmentation of textured and textureless objects through interactive
  perception.
\newblock 2012.

\bibitem{bohg2017interactive}
Jeannette Bohg, Karol Hausman, Bharath Sankaran, Oliver Brock, Danica Kragic,
  Stefan Schaal, and Gaurav~S Sukhatme.
\newblock Interactive perception: Leveraging action in perception and
  perception in action.
\newblock {\em IEEE Transactions on Robotics}, 33(6):1273--1291, 2017.

\bibitem{calli2015benchmarking}
Berk Calli, Aaron Walsman, Arjun Singh, Siddhartha Srinivasa, Pieter Abbeel,
  and Aaron~M Dollar.
\newblock Benchmarking in manipulation research: The ycb object and model set
  and benchmarking protocols.
\newblock {\em arXiv preprint arXiv:1502.03143}, 2015.

\bibitem{caselles2020sensory}
Hugo Caselles-Dupr{\'e}, Michael Garcia-Ortiz, and David Filliat.
\newblock On the sensory commutativity of action sequences for embodied agents.
\newblock {\em arXiv preprint arXiv:2002.05630}, 2020.

\bibitem{colombo2001development}
John Colombo.
\newblock The development of visual attention in infancy.
\newblock {\em Annual review of psychology}, 52(1):337--367, 2001.

\bibitem{eitel2020learning}
Andreas Eitel, Nico Hauff, and Wolfram Burgard.
\newblock Learning to singulate objects using a push proposal network.
\newblock In {\em Robotics Research}, pages 405--419. Springer, 2020.

\bibitem{everingham2015pascal}
Mark Everingham, SM~Ali Eslami, Luc Van~Gool, Christopher~KI Williams, John
  Winn, and Andrew Zisserman.
\newblock The pascal visual object classes challenge: A retrospective.
\newblock {\em International journal of computer vision}, 111(1):98--136, 2015.

\bibitem{fischer2015flownet}
Philipp Fischer, Alexey Dosovitskiy, Eddy Ilg, Philip H{\"a}usser, Caner
  Haz{\i}rba{\c{s}}, Vladimir Golkov, Patrick Van~der Smagt, Daniel Cremers,
  and Thomas Brox.
\newblock Flownet: Learning optical flow with convolutional networks.
\newblock {\em arXiv preprint arXiv:1504.06852}, 2015.

\bibitem{gibson2014ecological}
James~J Gibson.
\newblock {\em The ecological approach to visual perception: classic edition}.
\newblock Psychology Press, 2014.

\bibitem{goff2019building}
Leni K~Le Goff, Oussama Yaakoubi, Alexandre Coninx, and Stephane Doncieux.
\newblock Building an affordances map with interactive perception.
\newblock {\em arXiv preprint arXiv:1903.04413}, 2019.

\bibitem{gupta2012using}
Megha Gupta and Gaurav~S Sukhatme.
\newblock Using manipulation primitives for brick sorting in clutter.
\newblock In {\em 2012 IEEE International Conference on Robotics and
  Automation}, pages 3883--3889. IEEE, 2012.

\bibitem{ha2018recurrent}
David Ha and J{\"u}rgen Schmidhuber.
\newblock Recurrent world models facilitate policy evolution.
\newblock In {\em Advances in Neural Information Processing Systems}, pages
  2450--2462, 2018.

\bibitem{hafner2020mastering}
Danijar Hafner, Timothy Lillicrap, Mohammad Norouzi, and Jimmy Ba.
\newblock Mastering atari with discrete world models, 2020.

\bibitem{he2017mask}
Kaiming He, Georgia Gkioxari, Piotr Doll{\'a}r, and Ross Girshick.
\newblock Mask r-cnn.
\newblock In {\em Proceedings of the IEEE international conference on computer
  vision}, pages 2961--2969, 2017.

\bibitem{johnson2010infants}
Scott~P Johnson.
\newblock How infants learn about the visual world.
\newblock {\em Cognitive Science}, 34(7):1158--1184, 2010.

\bibitem{kaufmann1995development}
Franz Kaufmann.
\newblock Development of motion perception in early infancy.
\newblock {\em European Journal of Pediatrics}, 154(4):S48--S53, 1995.

\bibitem{lin2014microsoft}
Tsung-Yi Lin, Michael Maire, Serge Belongie, James Hays, Pietro Perona, Deva
  Ramanan, Piotr Doll{\'a}r, and C~Lawrence Zitnick.
\newblock Microsoft coco: Common objects in context.
\newblock In {\em European conference on computer vision}, pages 740--755.
  Springer, 2014.

\bibitem{lyubova2016passive}
Natalia Lyubova, Serena Ivaldi, and David Filliat.
\newblock From passive to interactive object learning and recognition through
  self-identification on a humanoid robot.
\newblock {\em Autonomous Robots}, 40(1):33--57, 2016.

\bibitem{oh2019video}
Seoung~Wug Oh, Joon-Young Lee, Ning Xu, and Seon~Joo Kim.
\newblock Video object segmentation using space-time memory networks.
\newblock In {\em Proceedings of the IEEE International Conference on Computer
  Vision}, pages 9226--9235, 2019.

\bibitem{philipona2008developpement}
David Philipona.
\newblock D{\'e}veloppement d'un cadre math{\'e}matique pour une th{\'e}orie
  sensorimotrice de l'exp{\'e}rience sensorielle.
\newblock 2008.

\bibitem{philipona2003there}
David Philipona, J~Kevin O'Regan, and J-P Nadal.
\newblock Is there something out there? inferring space from sensorimotor
  dependencies.
\newblock {\em Neural computation}, 15(9):2029--2049, 2003.

\bibitem{philipona2004perception}
David Philipona, Jk~O'regan, J-P Nadal, and Olivier Coenen.
\newblock Perception of the structure of the physical world using unknown
  multimodal sensors and effectors.
\newblock In {\em Advances in neural information processing systems}, pages
  945--952, 2004.

\bibitem{redmon2016you}
Joseph Redmon, Santosh Divvala, Ross Girshick, and Ali Farhadi.
\newblock You only look once: Unified, real-time object detection.
\newblock In {\em Proceedings of the IEEE conference on computer vision and
  pattern recognition}, pages 779--788, 2016.

\bibitem{schiebener2014physical}
David Schiebener, Ale{\v{s}} Ude, and Tamim Asfour.
\newblock Physical interaction for segmentation of unknown textured and
  non-textured rigid objects.
\newblock In {\em 2014 IEEE International Conference on Robotics and Automation
  (ICRA)}, pages 4959--4966. IEEE, 2014.

\bibitem{schiebener2011segmentation}
David Schiebener, Ale{\v{s}} Ude, Jun Morimoto, Tamim Asfour, and R{\"u}diger
  Dillmann.
\newblock Segmentation and learning of unknown objects through physical
  interaction.
\newblock In {\em 2011 11th IEEE-RAS International Conference on Humanoid
  Robots}, pages 500--506. IEEE, 2011.

\bibitem{shenigibson}
Bokui Shen*, Fei Xia*, Chengshu Li*, Roberto Mart{\'i}n-Mart{\'i}n*, Linxi Fan,
  Guanzhi Wang, Shyamal Buch, Claudia D’Arpino, Sanjana Srivastava, Lyne~P
  Tchapmi, Kent Vainio, Li~Fei-Fei, and Silvio Savarese.
\newblock igibson, a simulation environment for interactive tasks in large
  realistic scenes.
\newblock {\em arXiv preprint arXiv:2012.02924}, 2020.

\bibitem{teed2020raft}
Zachary Teed and Jia Deng.
\newblock Raft: Recurrent all-pairs field transforms for optical flow.
\newblock {\em arXiv preprint arXiv:2003.12039}, 2020.

\bibitem{wise2016fetch}
Melonee Wise, Michael Ferguson, Derek King, Eric Diehr, and David Dymesich.
\newblock Fetch and freight: Standard platforms for service robot applications.
\newblock 2016.

\bibitem{xu2015autoscanning}
Kai Xu, Hui Huang, Yifei Shi, Hao Li, Pinxin Long, Jianong Caichen, Wei Sun,
  and Baoquan Chen.
\newblock Autoscanning for coupled scene reconstruction and proactive object
  analysis.
\newblock {\em ACM Transactions on Graphics (TOG)}, 34(6):1--14, 2015.

\end{thebibliography}
\bibliographystyle{plain}

\clearpage
\appendix

\section{Flownet vs RAFT mask prediction qualitative comparison}
\label{app:raft_vs_flownet}

\begin{figure}[h]
    \centering
    \includegraphics[scale=0.38]{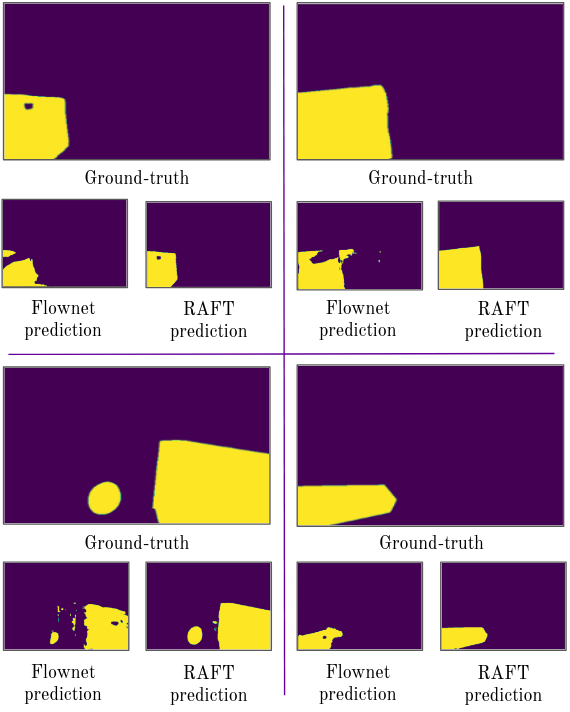}
    \caption{Comparison of mask predictions between Flownet and RAFT. While Flownet is able to roughly predict the object mask, RAFT has a better accuracy.}
    \label{fig:flownet_vs_raft}
\end{figure}

In Fig.\ref{fig:flownet_vs_raft}, we provide a qualitative comparison between object mask prediction of Flownet and RAFT. Both models are able to detect the moved object, hence they are suited for the SCOD method. However, the quantitative gap shown in Tab.\ref{tab:quanti-res} is explained by the higher accuracy of RAFT over Flownet. The exact shape of the objects is not perfectly captured by Flownet. On the contrary, RAFT is often able to predict the exact shape of the moved objects.

\section{Object tracking results}
\label{app:obj_tracking}

\begin{figure}[h]
    \centering
    \includegraphics[scale=0.3]{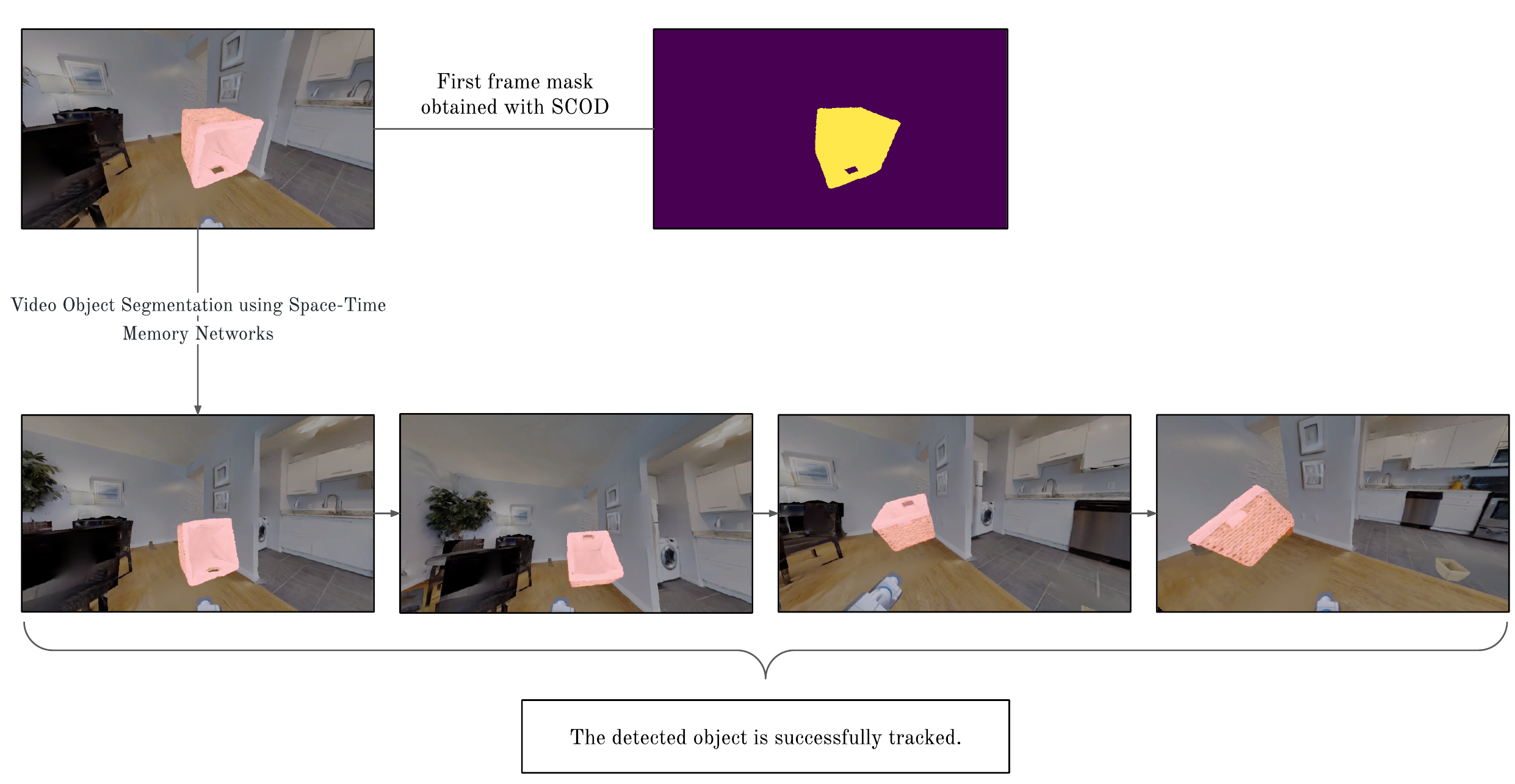}
    \caption{Object detection and tracking pipeline. We first use SCOD to detect an object, and use the learned mask to track it using STM, a semi-supervised video object segmentation algorithm.}
    \label{fig:tracking_res}
\end{figure}

Fig.\ref{fig:tracking_res} illustrates the detection + tracking pipeline using SCOD and STM \cite{oh2019video}. Video illustrations are also available in the supplementary video joint to the paper\footnote{\url{https://youtu.be/Bc5fwZH-CQU}}.

\section{Algorithm design alternatives} 
\label{app:design}
\begin{figure}[h!]
    \centering
    \includegraphics[scale=0.8]{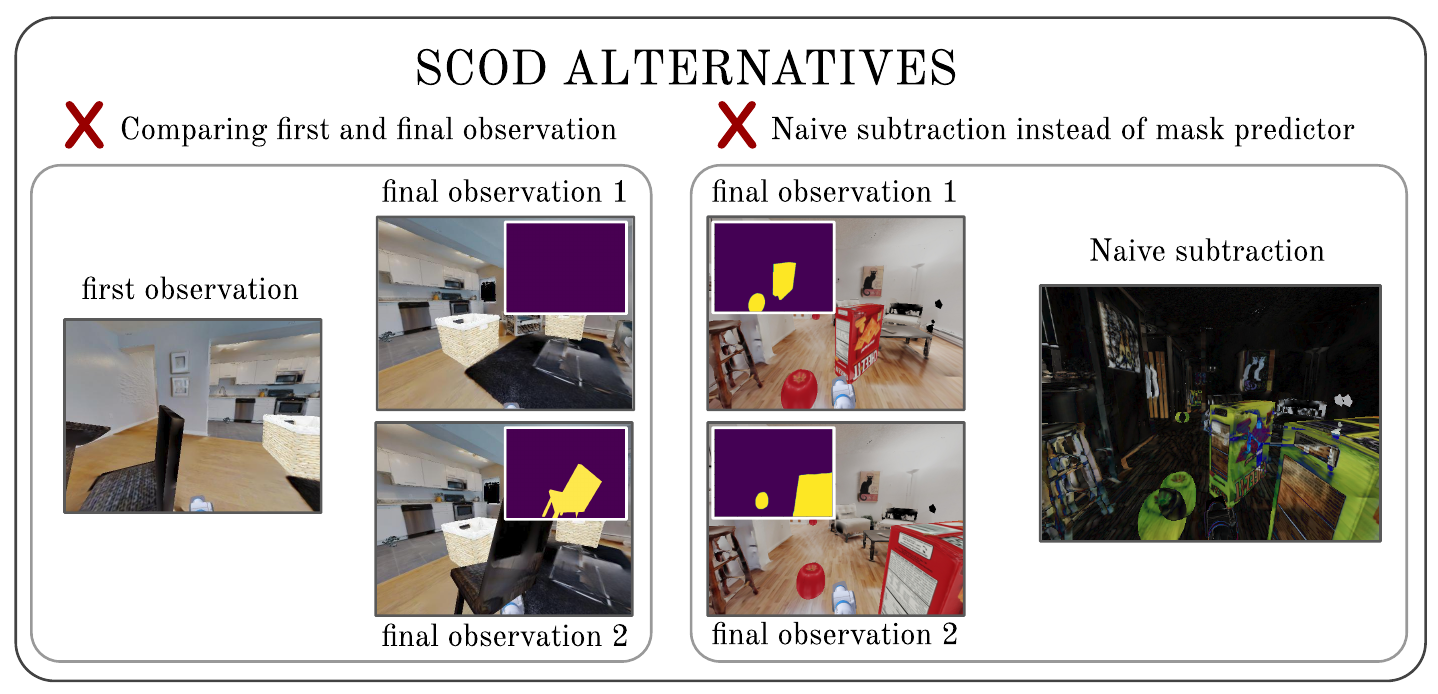}
    \caption{Algorithm design alternatives for SCOD. Comparing first and last observations fails, or replacing the mask predictor by naive subtraction are not viable options, which justifies the use of SC-experiments and mask predictor in SCOD. For comparison we provide the masks computed by SCOD over the final observations.}
    \label{fig:alternatives}
\end{figure}

We question design choices in the SCOD algorithm. First, we question the use of SC-experiments compared to simpler alternatives such as just playing an action sequence and comparing the first and last observations. This method would rarely detect objects because most experiments would result in a complete image change where the SC-experiments would highlight only a particular object, as illustrated in Fig.\ref{fig:alternatives}.
Another alternative would be to start in a position, play an action sequence, and then go back to this starting point and compare what's changed. While this approach would be comparable for movable object detection, this would not allow detecting immovable objects and free-space. 

Second, an alternative to the use of a mask predictor would be to use a naive image subtraction between the two observations resulting from the SC-experiment. However, if one object has moved, the naive image subtraction results in two masks (one for each position of the object that has moved). These masks can overlap, and thus be hard to distinguish. Then if more than one object has moved, the subtraction will prove difficult to interpret. This issue is illustrated in Fig.\ref{fig:alternatives}. We initially tried this, and then switched to the mask predictor which ended up being more efficient.

\section{Real robot results illustration}
\label{app:real-life}

\begin{figure}[h!]
    \centering
    \includegraphics[scale=1]{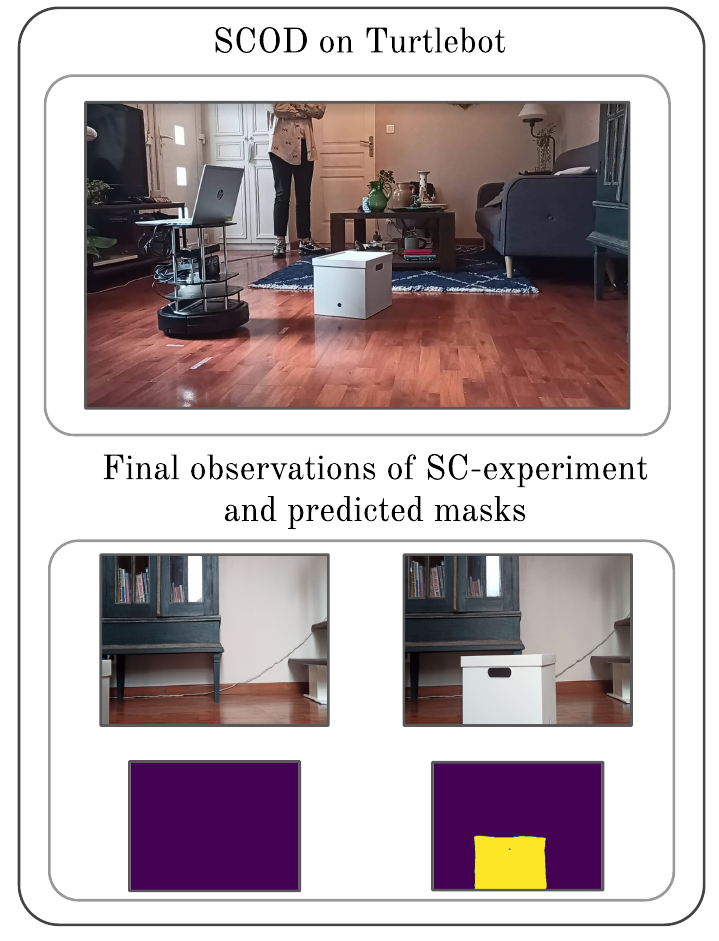}
    \caption{Object detection on Turtlebot using SCOD: the robot performed a SC-experiment which led to a moved object that is detected by SCOD, see supplementary video. We present the two resulting observations from the SC-experiment, and the predicted masks below. The algorithm has solely been trained on synthetic images, and generalizes to real-life scenarios.}
    \label{fig:real-life}
\end{figure}

\end{document}